\documentclass[a4paper,fleqn]{cas-sc}

\usepackage[authoryear,longnamesfirst]{natbib}

\def\tsc#1{\csdef{#1}{\textsc{\lowercase{#1}}\xspace}}
\tsc{WGM}
\tsc{QE}
\tsc{EP}
\tsc{PMS}
\tsc{BEC}
\tsc{DE}

\begin{document}
\let\WriteBookmarks\relax
\def\floatpagepagefraction{1}
\def\textpagefraction{.001}

\shorttitle{Knowledge-Aware Audio-Grounded Generative Slot Filling for Limited Annotated Data}
\shortauthors{Guangzhi Sun et~al.}

\title [mode = title]{Knowledge-Aware Audio-Grounded Generative Slot Filling for Limited Annotated Data}                       



%
\author[1]{Guangzhi Sun}


\fnmark[1]

\ead{gs534@cam.ac.uk}

\affiliation[1]{organization={University of Cambridge Department of Engineering},
    addressline={Trumpington Street}, 
    city={Cambridge},
    postcode={CB2 1PZ}, 
    country={United Kingdom}}

\author[1, 3]{Chao Zhang}

\ead{cz277@tsinghua.edu.cn}

\author[1, 2]{Ivan Vuli{\'c}}

\ead{ivan@poly-ai.com}

\author[2]{Pawe{\l} Budzianowski}

\ead{pawel@poly-ai.com}

\author[1]{Philip C. Woodland}

\ead{pw117@cam.ac.uk}

\cormark[1]

\affiliation[2]{organization={PolyAI Ltd.},
    address={10 York Rd},
    city={London},
    postcode={SE1 7ND}, 
    country={United Kingdom}}

\affiliation[3]{organization={Tsinghua University},
    address={30 Shuangqing Rd},
    city={Beijing},
    postcode={100190}, 
    country={China}}

\cortext[cor1]{Corresponding author}

\fntext[fn1]{Supported by a Cambridge International Scholarship from the Cambridge Commonwealth, European \& International Trust}


\begin{abstract}
Manually annotating fine-grained slot-value labels for task-oriented dialogue (ToD) systems is an expensive and time-consuming endeavour. This motivates research into slot-filling methods that operate with limited amounts of labelled data. Moreover, the majority of current work on ToD is based solely on text as the input modality, neglecting the additional challenges of imperfect automatic speech recognition (ASR) when working with spoken language. In this work, we propose a {K}nowledge-{A}ware {A}udio-{G}rounded generative slot filling framework, termed KA2G, that focuses on few-shot and zero-shot slot filling for ToD with speech input. KA2G achieves robust and data-efficient slot filling for speech-based ToD by {1)} framing it as a text generation task, {2)} grounding text generation additionally in the audio modality, and {3)} conditioning on available external knowledge (\textit{e.g.} a predefined list of possible slot values). We show that combining both modalities within the KA2G framework improves the robustness against ASR errors. Further, the knowledge-aware slot-value generator in KA2G, implemented via a pointer generator mechanism, particularly benefits few-shot and zero-shot learning. Experiments, conducted on the standard speech-based single-turn SLURP dataset and a multi-turn dataset extracted from a commercial ToD system, display strong and consistent gains over prior work, especially in few-shot and zero-shot setups.
\end{abstract}


\begin{highlights}
\item A knowledge-aware audio-grounded (KA2G) generative slot-filling framework is proposed for use with limited annotated data for task-oriented dialogue (ToD). KA2G formulates slot filling as a language generation task with natural language prompts. Slot value generation (SVG) is also grounded based on speech input via an ASR module in addition to a pre-trained language model (PLM) in order to give robustness to recognition errors.
\item KA2G integrates external contextual knowledge by using two tree-constrained pointer generator (TCPGen) components, one for ASR and one for SVG, with shared prefix-tree encoding networks. The use of TCPGen greatly benefits the KA2G slot-filling performance, especially on rare and unseen entities and unseen slot types.
\item Experiments on the SLURP dataset with speech input showed that the proposed KA2G framework can produce state-of-the-art slot-filling results. KA2G was further evaluated on an in-house multi-turn ToD dataset, CONCIERGE, to validate the effectiveness of KA2G in real-world applications.
\item Large and consistent improvements with the full KA2G framework were obtained over a standard pipeline-based ToD baseline on both datasets. The improvements were most prominent for rare and unseen entities on both datasets, with a 4.6\% absolute SLU-F1 increase for few-shot entities, an 11.2\% increase for zero-shot entities and 13.6\% increase for unseen slot types in SLURP. Meanwhile, KA2G improved more than 20 joint goal accuracy (JGA) points in multi-turn evaluation on CONCIERGE. The importance and contributions of the two TCPGen components were verified in a series of ablation studies and other analyses.

\item The proposed KA2G framework has a number of key differences in the use of TCPGen in ToD. Compared to the previous conference paper \citep{TCPGenSLU}
\begin{itemize}
    \item Rather than formulating SLU as a sequence-tagging problem which is an audio-grounded extension of text-based methods that are only able to handle predefined slot types, KA2G handles slot-filling as a generative task that depends on both audio input and a knowledge base. It is able to handle an open set of slot types using natural language queries and generates natural language slot values. We also demonstrate that KA2G is more robust to ASR errors and achieves better performance.
    \item While TCPGen and structured knowledge were mainly considered from the ASR perspective in \citet{TCPGenSLU}, in KA2G, a stacked TCPGen structure with a shared tree encoding network is adopted. Notably, TCPGen is applied to SVG to guide the generation process using the most relevant knowledge from multiple ASR alternatives.
\end{itemize}
\end{highlights}

\begin{keywords}
slot filling \sep spoken language understanding \sep audio-grounding \sep contextual biasing \sep knowledge base \sep generative model \sep limited data \sep few-shot \sep zero-shot
\end{keywords}

\maketitle

\section{Introduction}
\textit{Slot filling}, as a crucial natural language understanding component of task-oriented dialogue (ToD) systems, aims at filling in the correct value for predefined slots (\textit{e.g.} restaurant and hotel names) \citep{Tur:2010atis,tur2011spoken}. As manual fine-grained annotation for slot labels is expensive, time-consuming, and usually requires domain expertise \citep{Casanueva:2022nlupp}, increasing demands have been put on the performance of slot-filling systems under \textit{few-shot} or even \textit{zero-shot} learning setups \citep{fewshotsf1,fewshotsf2,BERT4SFlowdata}. Following the now-prevalent use of large Transformer-based pretrained language models (PLM) \citep{GPT2,BERT,T5} for transfer learning across many NLP tasks, PLMs have also been widely adopted in ToD for slot-filling tasks with limited labelled data \citep{BERT4SF,BERT4SFlowdata}.

More recently, other text-based approaches have reformulated slot filling as a question-answering (QA) or a sequence generation task, in order to further exploit the power of QA-oriented and generative PLMs \citep{QASF1,QASF2,SimpleTOD,QASF4}, especially in low-data scenarios \citep{QASF3,QASFzero,MultiwozZero2}
However, all these approaches operate directly on `perfect' text input, thus overestimating the performance of speech-based ToD systems where a loss in performance might occur due to imperfect automatic speech recognition (ASR) \citep{Gerz:2021minds}. Imperfect ASR output can especially harm slot filling that deals with entities infrequent in the general language (e.g., atypical personal, restaurant or hotel names) and is even more pronounced in situations with limited annotated data.

While very recent research has started to explore end-to-end slot-filling tasks for ToD with speech input \citep{SLU1,SLU2}, in this work we focus on a particularly challenging situation which is typically met in production: limited annotated data with many rare entities. Therefore, we propose KA2G, a Knowledge-Aware Audio-Grounded generative slot-filling framework which is tailored towards improving the robustness and performance of slot filling with spoken input. 

KA2G integrates information from both audio and text as input to a slot-value generator (SVG) which then generates textual fillers for each slot. Note that the final generation is also grounded in the audio modality. This mitigates the issues arising from noisy ASR-generated transcriptions. KA2G particularly boosts the performance of rare and unseen entities by learning to exploit the available external knowledge (e.g., predefined lists of possible values for slots) via two tree-constrained pointer generator (TCPGen) components \citep{TCPGen1,TCPGen2}. TCPGen builds a neural shortcut between the \textit{biasing list}, which is a list of entities likely to appear in a given context, and the model output via a pointer generator. Biasing lists are extracted from an external knowledge base (KB) containing possible entities for each slot type and are structured as subword-based prefix trees to be searched.
The first TCPGen is applied on the ASR side
to reduce ASR errors on high-value biasing entities based on the available context.  
The second TCPGen is applied on the SVG side 
to bias the generator's output using sub-trees which contain branches on the prefix-trees that are traversed during ASR beam search. The entire KA2G model is jointly optimised in an end-to-end fashion from the input-speech-`end' to the generated slot-value-`end'. The code for KA2G is available at \url{https://github.com/the-anonymous-bs/espnet/tree/master/egs/slurp/asr1}

Although our previous conference paper \citep{TCPGenSLU} explored TCPGen in SLU tasks, the proposed KA2G framework is fundamentally different in the following two key aspects:
\begin{itemize}
    \item Rather than formulating SLU as a sequence-tagging problem which is an audio-grounded extension of the text-based methods that are only able to handle predefined slot types, KA2G handles slot-filling as a generative task that depends on both audio input and a knowledge base. It handles an open set of slot types using natural language queries and generates natural language slot values. We also demonstrate that KA2G is more robust to ASR errors and achieves better performance.
    
    \item While TCPGen and structured knowledge were mainly considered from the ASR perspective in \citet{TCPGenSLU}, in KA2G, a stacked TCPGen structure with a shared tree encoding network is adopted here. Notably, TCPGen was applied to SVG to guide the generation process using the most relevant knowledge from multiple ASR alternatives.
\end{itemize}

The main experiments were conducted on two structurally different datasets with speech input, with a focus on few-shot and zero-shot learning scenarios: {1)} the single-turn SLURP dataset \citep{slurp}, and {2)} an in-house multi-turn ToD dataset extracted from a commercial concierge/booking system (henceforth termed CONCIERGE). 
While the zero-shot setup stretches the abilities of the tested systems to the extreme, the few-shot learning scenario is more pragmatic and suitable for industry research \citep{Lauscher:2020emnlp,BERT4SFlowdata}, as a small number of labels for each entity can usually be made available. 
Large and consistent improvements with the full KA2G framework were found over a standard pipeline-based ToD baseline on both datasets, \textit{e.g.}, improving by more than 20 joint goal accuracy (JGA) points in multi-turn evaluations on CONCIERGE. The improvements were most prominent for rare and unseen entities on both datasets. The importance and contributions of the two TCPGen components were verified in a series of ablation studies and other analyses. 

The rest of this article is organised as follows. Section \ref{sec:relwork} reviews related studies. Section \ref{sec:sf} introduces the KA2G framework, with a detailed explanation of TCPGen and how it can be applied to slot-filling. Section \ref{sec:setup} describes the experimental setup, and Section \ref{sec:results} discusses the results. Finally, conclusions are provided in Section \ref{sec:conclusion}.

\section{Related Work}
\label{sec:relwork} 
\subsection{Slot Filling as a Text Generation Task} 
Recent research has seen increased interest in reformulating the slot-filling task beyond standard sequence labelling and classification paradigms \cite{Shah:2019robust,Budzianowski:2019hello,BERT4SF,Coope:2020acl}. \citet{QASF1} and \citet{QASF2} recast slot filling as a QA task and a reading comprehension task, respectively, with both studies focusing on applications with limited data. More recently, \citet{QASF3} performed a comprehensive analysis of the QA approach for slot filling and provided both efficient and effective fine-tuning methods for domain-specific slot-filling models. 

Formulating slot filling as a text generation task has recently also become an active research area. \citet{QASF5} proposed a generative slot-filling framework that leverages PLMs fine-tuned on specific tasks and domains to improve task-/domain-specific generation. Another research stream focused on framing dialogue state tracking (DST) for multi-turn ToD as a text generation task. In particular, the T5DST model \citep{MultiwozZero2} utilised different slot descriptions as the prompt for generation for cross-domain DST. However, previous approaches have only dealt with text input and do not use external knowledge, whereas our proposed KA2G framework is audio-grounded and efficiently leverages external knowledge.

\subsection{Knowledge Integration for Slot Filling}
Research has also been performed on leveraging external knowledge bases or the ontology of a dialogue system for slot-filling. 
In \citet{schema}, domain-slot relations from the dialogue ontology were encoded using a graph neural network (GNN) to guide the system, while \citet{KAGEDST} further extended the use of the GNNs to capture correlations between slots and values in different domains. For slot filling for ToD with speech input, \citet{speech2slot} used a Transformer encoder to encode external knowledge into hidden representations, while \citet{TCPGenSLU} built a neural shortcut from the external knowledge base directly to the slot filling output. While both methods focused on zero-shot learning setups, they relied on the standard sequence labelling formulation of the slot-filling task. In contrast, KA2G adopts a more flexible generative framework, which yields improved performance in few-shot and zero-shot scenarios. 

\subsection{Contextual Knowledge Integration in ASR} 
Previous studies on contextual biasing have been focused on either shallow-fusion-based score-level interpolation \citep{shallow_context_1,shallow_context_2,shallow_context_3} or deep neural encoders or representations \citep{biasing1,biasing2,biasing3,biasing4}. Recent work also explored the combination of deep and shallow approaches for contextual biasing. Specifically, \cite{deepshallow,DBRNNT} proposed to apply shallow fusion and deep biasing together in the end-to-end ASR model. More recently, pointer-generator-style shortcuts \cite{TCPGen1, biasing5} or neural-FST \citep{biasing6} approaches that directly modify the final ASR output distribution have been investigated which allowed joint optimisation of the entire network in an end-to-end fashion. Meanwhile, TCPGen \cite{TCPGen1} also achieved high efficiency by using a symbolic prefix-tree search to handle biasing lists of thousands of words. Further research into TCPGen \citep{TCPGen2} used a graph neural network (GNN) to encode the prefix tree in TCPGen, which achieved further improvements in the recognition accuracy of biasing words. TCPGen with powerful GNN encodings acts as the backbone in both our previous work \citep{TCPGenSLU} and the proposed KA2G framework.

\section{Methodology}
\label{sec:sf}
\begin{figure*}[t]
\centering
    \includegraphics[width=0.95\textwidth]{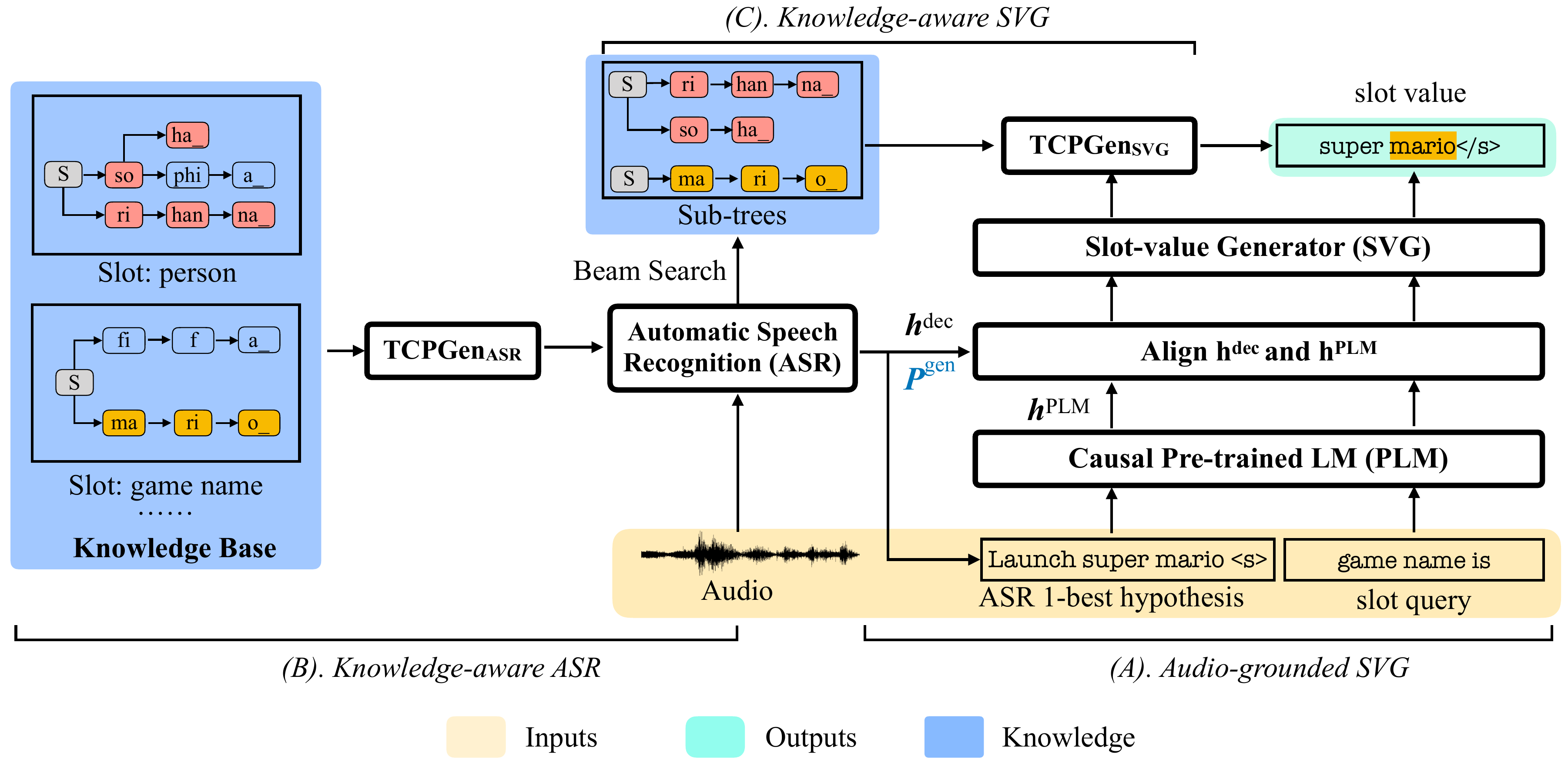}
    \caption{The KA2G framework for slot filling. Its three key components are indicated by the labels (A), (B) and (C) and are described in Section 3. Knowledge base showing two example slot types (person and game name) containing possible values structured as wordpiece prefix trees. The example in the figure shows that the first pointer generator network (TCPGen$_\text{ASR}$) traverses branches of \texttt{mario}, \texttt{soha} and \texttt{rihanna}, which are then included in sub-trees. Another generator network (TCPGen$_\text{SVG}$) uses this branch further to generate slot values.}
    \label{fig:model}
\end{figure*}

The KA2G framework is illustrated in Fig. \ref{fig:model}. It comprises three key components as follows:

\vspace{0.2cm}
\noindent \textit{(A)} The audio-grounded SVG module combines output representations from the ASR module and the text-only PLM to generate values based on the slot prompt. The audio-grounded SVG module, as explained in Section \ref{sec:A}, acts as the foundation of KA2G where the two proposed TCPGen components for knowledge integration are added.

\vspace{0.2cm}
\noindent \textit{(B)} The knowledge-aware ASR component that integrates external knowledge into KA2G via the first TCPGen component (TCPGen$_\text{ASR}$). The TCPGen component and how it is integrated into the ASR module of KA2G are explained in Section \ref{sec:B}, together with the slot shortlist prediction mechanism dedicated to slot-filling tasks to obtain a more focused biasing list. 

\vspace{0.2cm}
\noindent \textit{(C)} The knowledge-aware SVG further integrates knowledge explored during the ASR beam search via the second TCPGen (TCPGen$_\text{SVG}$). TCPGen$_\text{SVG}$ extends the scope of TCPGen-based contextual knowledge integration from ASR tasks to any general natural language generation tasks will be explained in detail in Section \ref{sec:C}.

\subsection{Audio-Grounded SVG}
\label{sec:A}

The audio-grounded SVG module comprises (i) an ASR module, (ii) a causal/autoregressive PLM, (iii) an alignment module, and (iv) the SVG; This is illustrated in the right side of Fig. \ref{fig:model}. The SVG is implemented as a single-layer unidirectional LSTM which takes the concatenated vectors from the ASR module and the PLM as the representation of the context to make predictions for the value with a given slot query. The LSTM architecture is used for simplicity and increased stability in low-resource setups, and to avoid over-parameterisation since both ASR and PLM have complex model structures with millions of parameters. The ASR model is an attention-based encoder-decoder (AED) where the decoder hidden states, $\mathbf{h}^\text{dec}$, are sent to the SVG.

\begin{figure}[h]
    \centering
    \includegraphics[width=0.55\linewidth]{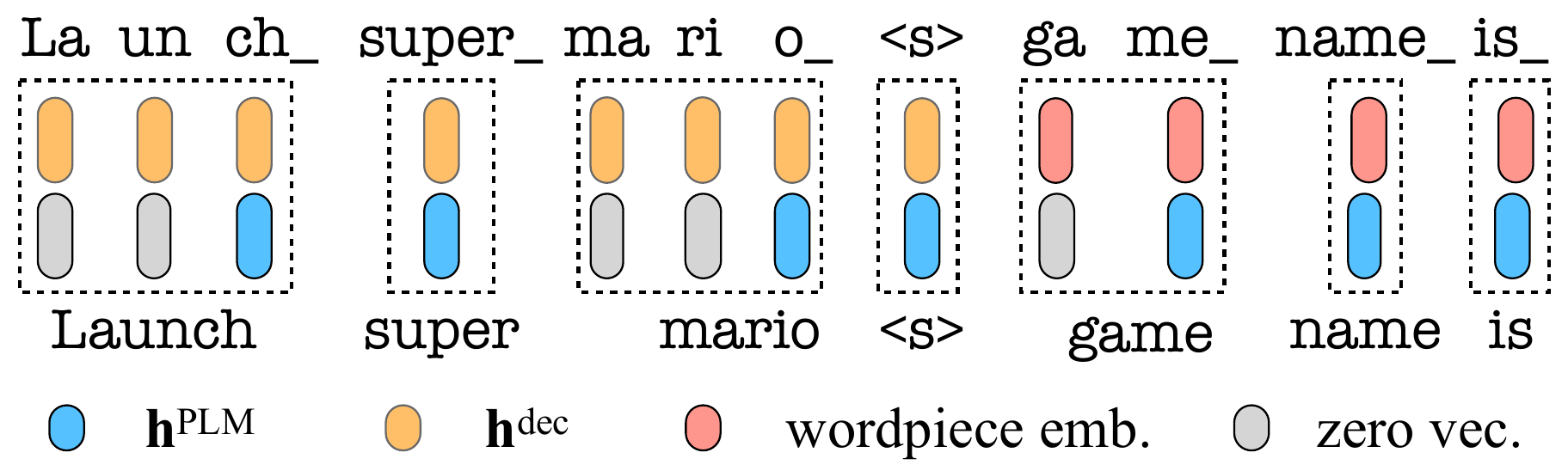}
    \caption{An example of aligning $\mathbf{h}^\text{PLM}$ and $\mathbf{h}^\text{dec}$.}
    \label{fig:align}
\end{figure}

As the label space for the PLM is too sparse to be used by the ASR module, the ASR module instead uses a much smaller subword token vocabulary than the PLM, and hence the sequence of ASR hidden states and the sequence of PLM output vectors $\mathbf{h}^\text{PLM}$ will be asynchronous. To resolve this \textit{(mis-)alignment issue}, the output of the SVG is first set to have the same subword tokens as the ASR module, which helps to make the best use of the acoustic information. 
The PLM outputs are then taken for each (full) word instead of every subword at each word end and aligned with $\mathbf{h}^\text{dec}$ at each word ending subword before concatenation. For non-terminal subwords, zero-vector padding is used as placeholders for the PLM output. 

An example of this alignment is shown in Fig. \ref{fig:align}. The alignment of $\mathbf{h}^\text{dec}$ and $\mathbf{h}^\text{PLM}$ is therefore achieved at word ends, and using the same subwords for both ASR and PLM is not necessary. Moreover, for a slot query which prompts the generation (e.g. \texttt{the person is} or \texttt{the game name is}), as there is no corresponding input audio, the embedding of the preceding wordpiece is used in place of $\mathbf{h}^\text{dec}$. Note that this alignment mechanism allows the slot value generation to use the PLM as well: Whenever a new word is generated, a new $\mathbf{h}^\text{PLM}$ with the new word can be obtained which is then concatenated with the preceding wordpiece to generate the next one.

The SVG is trained end-to-end by jointly optimising the ASR and slot-value generation criteria as shown below:
\begin{equation}
    \mathcal{L}_\text{joint} = \mathcal{L}_\text{ASR} + \mathcal{L}_\text{SVG},
    \label{eq:joint}
\end{equation}
where the respective sub-losses are defined as
\begin{align}
    \mathcal{L}_\text{ASR}&= \text{log}~P(\mathbf{y}_{1:n}|\mathbf{x}_{1:T}),\\
    \mathcal{L}_\text{SVG}&=\text{log}~P(\mathbf{s}_{1:m}|\mathbf{q}_{1:k},\mathbf{h}^\text{dec},\mathbf{h}^\text{PLM}).
    \label{eq:svg}
\end{align}
Here, $\mathbf{y}_{1:n}$ is the subword ASR sequence, $\mathbf{s}$ is the generated slot value sequence, $\mathbf{q}$ represents the query sequence (e.g. \texttt{the game name is}) and $\mathbf{x}_{1:T}$ is the sequence of acoustic features. Note that $\mathbf{h}^\text{PLM}$ in the SVG loss covers not only the context but also the slot query and value using the aforementioned alignment mechanism. In order to allow the model to also handle predictions for slots not present in the utterance, $N_n$ (randomly sampled) slots that are not mentioned in the context are incorporated in training as negative examples, where $N_n$ is a hyper-parameter. The model should learn to generate \texttt{None} values for those `not-present' slots. The entire SVG, together with the PLM and the two TCPGen components are optimised using the SVG loss.

During inference, $\mathbf{h}^\text{PLM}$ is obtained by feeding the 1-best ASR hypothesis to the PLM. The same context is prompted with all possible slot types, and those that do not output a \texttt{None} value are saved. For multi-turn ToD, the dialogue history is encoded by the PLM before the start of the current context.

\subsection{Knowledge-Aware ASR}
\label{sec:B}
External knowledge is organised as a dictionary of slots along with the possible values per slot: see the left blue block in Fig. \ref{fig:model}. 
It conditions the ASR via contextual biasing using TCPGen$_\text{ASR}$ (as shown in Fig. \ref{fig:model}).
Contextual biasing is an effective method to boost the recognition of rare words or entities in end-to-end trainable ASR systems, which represents the knowledge as a biasing list \citep{DBRNNT,biasing1,biasing2,biasing3,biasing4,biasing5,biasing6}. The biasing list contains a list of biasing entities that are likely to appear in a given context, such as a particular restaurant name or the name of an artist in a playlist, and the recognition accuracy can be improved if they are included in the biasing list. In slot filling, possible named entities for each slot type can be collected to form a structured KB, and the biasing list can be extracted from the KB as explained in Section \ref{sec:setup}.

\subsubsection{TCPGen}
\label{sec:tcpgen}

TCPGen \citep{TCPGen1} is a neural component combining symbolic prefix-tree search with a neural pointer generator for contextual biasing, which enables end-to-end optimisation with ASR systems. Although in this section, TCPGen is described based on TCPGen$_\text{ASR}$, TCPGen$_\text{SVG}$ which is presented later in Section 3.3, is also based on the same mechanism. At each output step of ASR, TCPGen$_\text{ASR}$ calculates a distribution over all valid subwords, referred to as the TCPGen$_\text{ASR}$ distribution, constrained by a subword-level prefix-tree built from the biasing list. TCPGen$_\text{ASR}$ also predicts a generation probability $P^\text{gen}$ indicating how much contextual biasing is needed at a specific step. If there are valid paths found in the tree, the set of valid subwords is copied to the ASR output by interpolating the TCPGen$_\text{ASR}$ distribution and the original ASR model output distribution, weighted by the generation probability.

\begin{figure*}[t]
    \centering
    \includegraphics[width=0.9\linewidth]{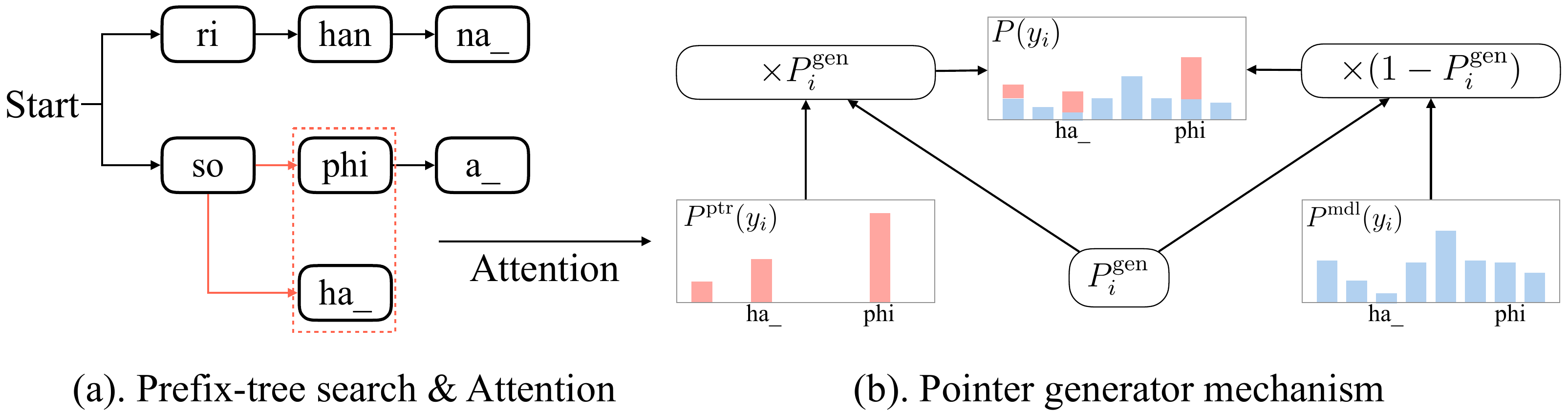}
    \caption{Illustration of TCPGen component for ASR (TCPGen$_\text{ASR}$) with corresponding terms in Eqn. \eqref{eq:TCPGen_final}. (a). If ``so" is the preceding token, ``phi" and ``ha\_" are two valid wordpieces with non-zero $P^\text{ptr}$. Note that TCPGen$_\text{SVG}$ introduced next is also based on this mechanism. (b). $P^\text{ptr}(y_i)$ is the TCPGen distribution. $P^\text{mdl}(y_i)$ is the distribution generated by a standard end-to-end model. $P(y_i)$ is the final output distribution. $P^\text{gen}_i$ is the generation probabilities.}
    \label{fig:tcpgen}
\end{figure*}

An illustration of the computation of TCPGen using the same example prefix tree as Fig. \ref{fig:model} is shown in Fig. \ref{fig:tcpgen}. During ASR decoding, a set of valid subwords is obtained by searching the prefix tree with the decoded preceding subwords. Then, scaled dot-product attention is performed to obtain the TCPGen$_\text{ASR}$ distribution $P^{\text{ptr}}(y_{i})$ (omitting dependencies on $y_{1:i-1}$ and $\mathbf{x}_{1:T}$ for presentation clarity) as follows:
\begin{equation}
    P^{\text{ptr}}(y_{i}) = \text{Softmax}(\text{Mask}({\mathbf{q}_i\mathbf{K}^\text{T}}/{\sqrt{d}})),
    \label{eq:TCPGen_attention}
\end{equation}
where $d$ is the dimensionality of $\mathbf{q}_i$ and Mask$(\cdot)$ sets the probabilities of subwords that do not form valid paths at the current step to zero. The query vector $\mathbf{q}_i$ is computed from the context vector and the previously decoded token embedding.  The key and value vectors are node encodings of corresponding subwords on the prefix tree. To enable lookahead functionality and obtain more powerful node representations, a graph convolutional network (GCN) \citep{GCN,TCPGen2} was used to encode the nodes on the tree. This node encoding can be done efficiently

The generation probability is calculated using the decoder hidden states and the weighted combination of node encodings from the attention mechanism. Then, the final output can be calculated as follows:
\begin{equation}
    P(y_i) = P^{\text{mdl}}(y_i)(1-{P}^\text{gen}_i) + P^{\text{ptr}}(y_i)P^\text{gen}_i,
    \label{eq:TCPGen_final}
\end{equation}
where $P^{\text{mdl}}(y_i)$ represents the output distribution from the standard end-to-end model, and $P^{\text{gen}}_i$ is the generation probability. 

\subsubsection{Slot Shortlist Prediction for TCPGen in SLU} 
For slot filling, possible entities for each slot are structured as prefix trees separately as shown in Fig. \ref{fig:model}.
In order to have a more focused biasing list, instead of using all slots, a shortlist of slots is predicted at the start of decoding for each word using a class language model (CLM) \citep{TCPGenSLU,biasing4,biasing6} which takes the decoded word-level history as input. The top $K$ slot types predicted by the CLM is used as contextual knowledge, where one TCPGen distribution is calculated for each slot type to model the joint distribution of slot type and wordpieces. The TCPGen distribution used for the pointer generator is obtained by marginalising with respect to the slot types, i.e. summing up probabilities of the same wordpiece in all shortlist slots, as shown in Eqn. \eqref{eq:clmtcpgen2}:
\begin{equation}
    P^{\text{ptr}}(y_i) = \sum_{s\in\mathcal{S}} P^{\text{ptr}}(s,y_i)
    \label{eq:clmtcpgen2}
\end{equation}
Note that the top $K$ slot list is updated with the current decoded word history when there is no valid path found on any of the prefix trees.

Moreover, as the generation probability $P^\text{gen}$ provides an indication of how much contextual biasing is needed to decode each subword token, it is concatenated with $\mathbf{h}^\text{dec}$ and sent to the SVG to further indicate where in the context the knowledge has been used. TCPGen$_\text{ASR}$ is jointly optimised with the SVG module.

\subsection{Knowledge-Aware SVG}
\label{sec:C}

Alternative hypotheses are an essential resource that can be obtained from the ASR system, especially for low-frequency named entities. To exploit the knowledge available in additional ASR hypotheses, the knowledge-aware SVG module is proposed here: it extracts branches on each prefix-tree during ASR beam-search decoding and forms sub-trees to be used by TCPGen$_\text{SVG}$ (as shown in Fig. \ref{fig:model}) to integrate knowledge into the SVG. In particular, as each prefix tree used in the ASR beam search decoding is searched, a valid path that leads from the root node to a leaf node will be saved, which corresponds to a valid named entity belonging to that slot type. After decoding, the lists of entities corresponding to the valid paths found for each slot are gathered and organised into prefix trees. These prefix trees are essentially sub-trees of the original prefix trees for each slot type. 

Next, sub-trees are encoded using the same GCN as used in Section \ref{sec:B} and are searched when generating slot values in the same way as with TCPGen$_\text{ASR}$. In contrast to TCPGen$_\text{ASR}$, for TCPGen$_\text{SVG}$, the query comes from the SVG hidden state at each decoding step, while the keys and values are taken from the node encodings on the sub-trees. In the example shown in Fig. \ref{fig:model}, the beam search traverses the entity ``rihanna" and ``soha" in the \texttt{person} slot, and hence the sub-tree of the \texttt{person} slot is constructed using these two entities. When prompting the system with the slot \texttt{person}, this sub-tree is subsequently used to generate values. 

In this manner, possible entities that are not covered by the 1-best hypothesis but are explored in the ASR beam search can be effectively used via the copy mechanism in the SVG. In addition to the benefit of exploring other hypotheses, TCPGen$_\text{SVG}$ on the generation side also improves the performance on rare and unseen entities even if they are correctly recognised, as the SVG might still be unable to pick them out due to insufficient training samples. TCPGen$_\text{SVG}$, as a pointer generator mechanism, enables the SVG to directly copy entities from the relevant knowledge that is also filtered by the ASR system, even if they are not seen in training.

\section{Experimental Setup}
\label{sec:setup}

\subsection{Training and Evaluation Data} 
Experiments were performed on two structurally different speech-based English ToD datasets as described below.

\vspace{0.2cm}
\noindent \textbf{SLURP} \citep{slurp} is a collection of 72K audio recordings of single-turn user interactions with a home assistant, annotated with scenarios, actions and entities. We ran experiments in two data setups. {1)} The official training, validation and test split were used and, following \citet{espnetslu}, synthesised audio was used for training. {2)} A simulated zero-shot setup following \citet{TCPGenSLU} was used: in training, we held out all utterances containing entities of five randomly selected `unseen' slots were held out and then the held-out utterances were used for testing. 

External knowledge was organised as a simple dictionary for SLURP, referred to as the knowledge base (KB), where keys were slot types and values were lists of possible named entities for that type. It was created for experimental purposes by gathering named entities that appeared in the entire SLURP data for each slot type (including train, validation and test sets), as a simulation of a real-world task environment. The average size of these lists was 106: the largest list was \texttt{person} which contained 872 entities, and the smallest list was \texttt{transport\_agency}, which only contained 2 entities.

\begin{figure}[h]
    \centering
    \includegraphics[width=0.85\linewidth]{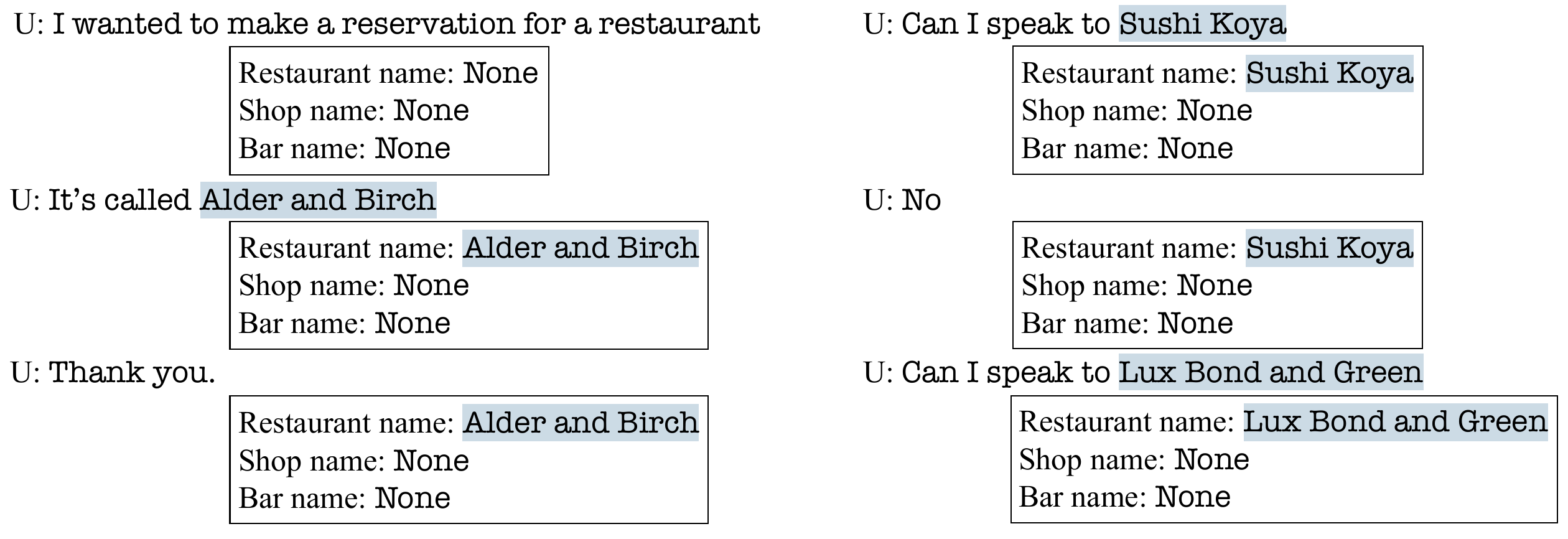}
    \caption{Two example dialogues from the CONCIERGE dataset along with the state tracking labels. Left: the slot filling label for the restaurant name in the last turn is \texttt{None}. Right: The user changed the goal in the third turn.}
    \label{fig:example}
\end{figure}

\vspace{0.2cm}
\noindent \textbf{CONCIERGE} is a multi-turn dialogue dataset obtained from a commercial system, and it represents a standard and challenging few-shot learning setup typically met in production. The dataset contains a collection of 8~kHz noisy phone-call conversations of real customer interactions with a concierge voice bot covering the \texttt{restaurant}, \texttt{shop} and \texttt{bar} domains. The audio was upsampled to 16~kHz to match the ASR model. Example dialogues from CONCIERGE are shown in Fig. \ref{fig:example}. Some dataset statistics are given in Table \ref{tab:polydata}. Each entity in the test set had only 1 to 5 occurrences in the training portion of the dataset and hence can be considered a `few-shot' scenario.

\begin{table}[h]
    \centering
    {
    \begin{tabular}{lccc}
    \toprule
     Split    &  \# Turns & \# Dialogues & Time (hours) \\
    \midrule
     Train    & 1934 & 829 & 3.17 \\
     Valid & 212 & 97 & 0.36 \\
     Test & 428 & 225 & 0.86 \\
     \bottomrule
    \end{tabular}
    }
    \caption{Statistics of the CONCIERGE dataset, including the number of dialogue turns (\# Turns), the number of dialogues (\# Dialogues) and the total time of speech (\# Time) in hours.}
    \label{tab:polydata}
\end{table}

\subsection{KA2G: Main Setup} 

The ASR model used an encoder with 16 Conformer blocks \citep{conformer} with a 512-dim hidden state and 4-head attention, a 1024-dim single-head location-sensitive attention and a 1024-dim single-layer unidirectional LSTM decoder. Each 10~ms frame of the input audio was represented by an 80-dim mel-scale filterbank feature. A suffix-based unigram WordPiece model \cite{unigramwp} with 600 distinct WordPieces was used for the output. 

GPT-2 \citep{GPT2} with a 768-dim output representation was used as the autoregressive PLM in all experiments. Both TCPGen components adopt one 256-dim single-head attention layer. A two-layer 256-dim GCN was used with the input node encodings set to the WordPiece embeddings of the ASR decoder, based on the default suggestions from prior work \cite{TCPGen2}. The dimensions of the slot-value generator and projection layers in TCPGen were all determined by the stated dimensionalities. The CLM used to predict a shortlist of slots for SLURP experiments was a single-layer 2048-dim LSTM. 

\subsection{Main Baseline}
The proposed KA2G framework was compared to a pipeline system which used the same ASR model to generate the 1-best hypothesis. The 1-best hypothesis was used as input text to the GPT-2 model for slot-value generation. As for KA2G, the GPT-2 model was also finetuned by generating slot values given slot prompts. In addition, the pipeline system can also be equipped with TCPGen$_\text{ASR}$, acting as a stronger baseline, which achieves better recognition accuracy on rare entities. The purpose of comparing to this system is to showcase the improvement attained by KA2G beyond just the improvement in recognition accuracy.

\subsection{KA2G: Training and Inference}

KA2G was implemented in ESPNet \citep{espnetslu}. The ASR AED models, together with the ASR-TCPGen component, were pretrained for 20 epochs on the Librispeech 960-hour English audiobook data \citep{TCPGen2}. 
When training the SVG (see Section 3.1), $N_n=10$ negative `not-present' slots were randomly chosen and added for each utterance in SLURP, and $N_n=3$ for CONCIERGE. The training was run on a single Nvidia A100 GPU.

Both the TCPGen$_\text{SVG}$ and TCPGen$_\text{ASR}$ components were trained in the same way as in \citep{TCPGen2}, where the full biasing list was first defined for each dataset, and the biasing list for each utterance was organised by selecting biasing entities in the reference transcription and adding a certain number of distractors following \citet{DBRNNT}.  For SLURP, the full biasing list was defined by selecting entities in the KB with fewer than 30 examples in the training set, including unseen entities. There were altogether 5,000 biasing entities. For TCPGen$_\text{ASR}$, the number of distractors was randomly picked between 100 and 200, whereas for TCPGen$_\text{SVG}$, 20 distractors from the same slot type were used, which was close to the size of the prefix-tree the model would see during in inference. A random drop of 30\% of the reference biasing entities was applied to both the TCPGen$_\text{ASR}$ and TCPGen$_\text{SVG}$. The same full biasing list selection criterion, as well as the training procedure for TCPGen, was also applied to the CONCIERGE data.

During inference, a beam size of 30 was used for ASR decoding. For SLURP, entities in the KB with fewer than 30 examples in training were used as biasing entities, and the CLM predicted the top 2 slot types at each word boundary. For experiments using the SLURP zero-shot learning split, a biasing list with 2k entities incorporated all entities in the unseen slots since they were all unseen entities, and was used as a whole during inference, as the CLM could not predict unseen slot types. For few-shot learning on CONCIERGE, all entities in the test set were included in the biasing list since they all appeared fewer than 5 times, which formed a biasing list of 105 entities. Since this was a much smaller biasing list, the entire list was used without CLM prediction. In the experiments, entities that appeared fewer than 5 times are referred to as \textit{few-shot entities}. For the CONCIERGE data, all biasing entities were few-shot entities. Unless stated otherwise, greedy decoding was used for SVG.

\subsection{Evaluation Metrics}

The ASR output is evaluated using the standard word error rate (WER) measure. For slot filling, the SLU-F1 \citep{slurp} and the micro Entity-F1 scores are used to measure performance, offering insights into both word-level and character-level F1 scores. Moreover, for multi-turn dialogues, joint goal accuracy (JGA) is reported: JGA counts a turn as correct only if all slots in the dialogue state are correctly filled in that turn. 

\section{Results and Discussion}
\label{sec:results}

\subsection{Experiments on SLURP}
\vspace{0.3cm}

This section describes the set of experiments performed on the SLURP dataset. It begins with a summary of the main results, followed by a discussion of key aspects. This includes a comparison to prior work, a discussion from the practical aspect about few-shot and zero-shot learning for slot filling, the ablation study, the impact of training data sizes, the impact of beam search, the impact of the incorporation of an external LM as well as the performance under the zero-shot setup.

\subsubsection{Main Results}
\begin{table*}[t]
    \centering
    {
    \begin{tabular}{lc|cccc}
    \toprule
    \textbf{System}  & \textbf{WER} (\%) $\downarrow$ & \textbf{Overall} F1 (\%) & 0 $<$ f $<$ 30 F1 (\%) & 0 $<$ f $<$ 5  F1 (\%) & f $=$ 0  F1 (\%)\\ 
    \cmidrule(lr){2-2} \cmidrule(lr){3-3} \cmidrule(lr){4-6}
     Pipeline    &  12.7 &  79.2 (73.0) & 72.4 (69.5) & 69.2 (69.1) & 21.1 (7.2) \\
     Audio-grounded SVG & 12.5 & 79.6 (73.9) & 74.3 (71.8) & 72.2 (71.9) & 20.9 (6.5) \\
     \cmidrule(lr){2-2} \cmidrule(lr){3-3} \cmidrule(lr){4-6}
     Pipeline + Contextual ASR   & 12.4 & 79.7 (73.7) & 73.5 (70.2) & 70.8 (70.5) & 24.5 (9.8) \\
     Full KA2G & \textbf{12.1} & \textbf{80.6 (75.1)} & \textbf{75.7 (73.4)} & \textbf{73.8 (73.7)} & \textbf{32.3 (18.6)}\\
     \bottomrule
    \end{tabular}
    }%
    \caption{WER and F1 scores, including SLU-F1, and Entity-F1 (in parentheses) on SLURP. F1 scores were measured for all entities (\textit{Overall}), as well as for biasing entities (occurrence frequency $f<30$), few-shot entities ($f<5$) and unseen entities ($f=0$). \textit{Pipeline} represented the pipeline system using the same AED-based ASR model to get the 1-best hypothesis, followed by GPT-2 for the slot-value generation. \textit{Contextual ASR} uses TCPGen$_\text{ASR}$ in the pipeline system.}
    \vspace{0.3cm}
    \label{tab:slurpresult1}
\end{table*}

The results on SLURP are summarised in Table \ref{tab:slurpresult1}. Both the SLU-F1 and Entity-F1 scores reveal that, compared to the pipeline system, using the audio-grounded SVG model (Row 2 in the table) achieved better performance overall, with much better F1 scores on biasing entities (\textit{i.e.}, their occurrence frequency was $0<f<30$) and few-shot entities ($0<f<5$). Similar improvements from audio-grounding were observed when comparing the full KA2G framework to the pipeline system with a contextual ASR using TCPGen. Overall, KA2G achieved a 1.4\% absolute SLU-F1 increase compared to the baseline pipeline system, with a 4.6\% SLU-F1 improvement on few-shot entities and an 11.2\% SLU-F1 improvement on unseen entities. This indicates that audio grounding helped the system to better deal with few-shot entities as long as at least some examples were exposed to it. Similar trends but large improvements were found in Entity-F1 than in SLU-F1 using KA2G, especially on entities in the biasing list as TCPGen is able to guide the slot-value generation to complete the entire entity correctly by following a specific path on the prefix-tree. Furthermore, similar to \cite{SLU2}, the model was also able to generate entities that were incorrectly recognised by ASR: for the audio-grounded system, 10\% of the correctly filled entities were not correctly recognised by the ASR system, whereas that ratio was only 3\% for the pipeline system. 

\begin{figure}[h]
    \centering
    \includegraphics[width=0.5\linewidth]{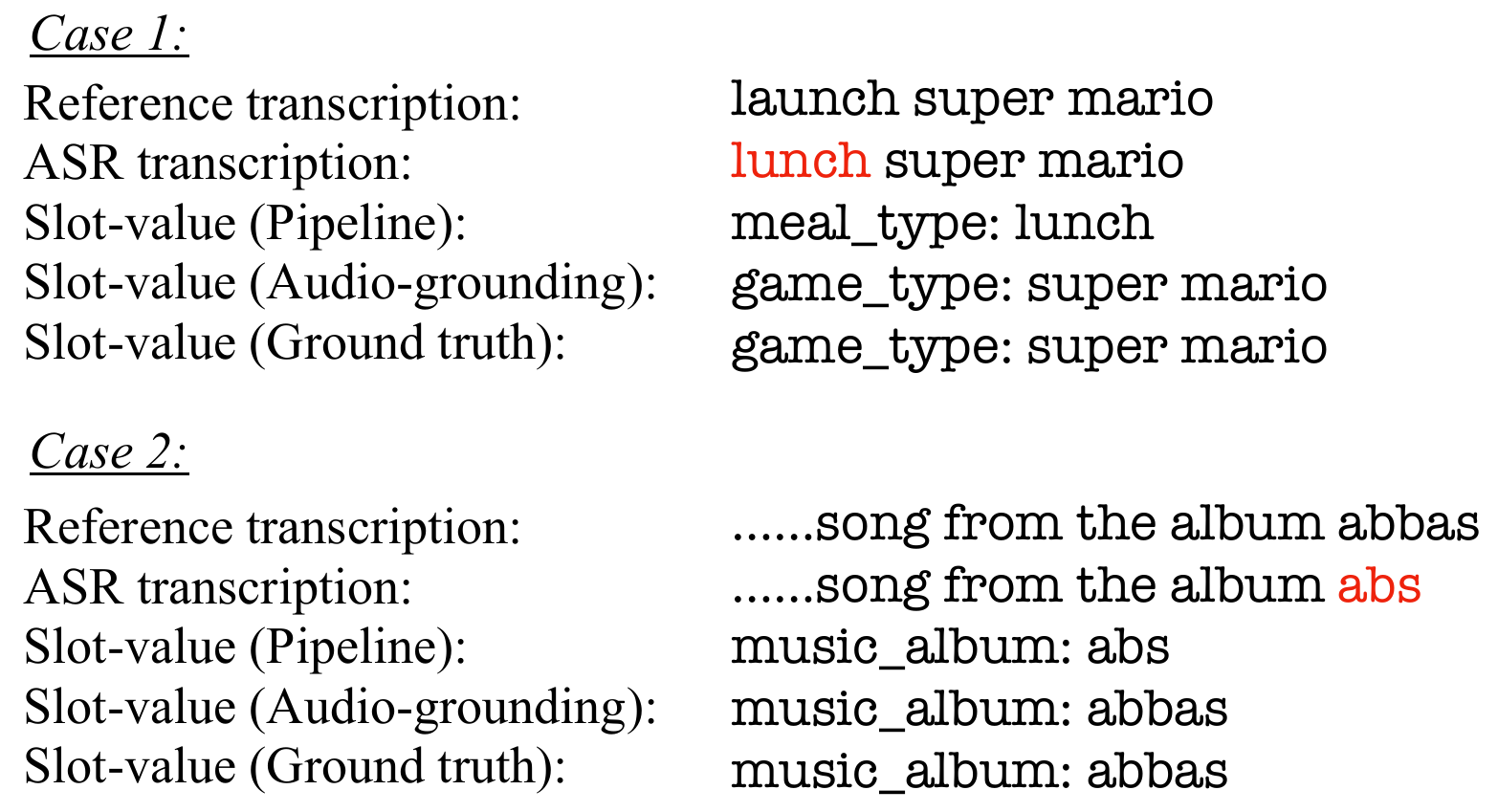}
    \caption{Two examples of utterances from SLURP where audio-grounding helped slot filling.}
    \label{fig:casestudy}
\end{figure}

In order to further investigate the influence of audio-grounding in slot filling, a case study was performed on SLURP. Two examples where the pipeline system and audio-grounded SVG had the same ASR output are shown in Fig. \ref{fig:casestudy}. In Case 1, while the semantic meanings of ``launch'' and ``lunch'' were completely different, their pronunciations were similar. Therefore, the ASR error ``lunch'' confused the pipeline system which produced the wrong slot type and value, which was not recoverable in the SVG-end since the audio information was lost in the pipeline system. However, with audio-grounding, the pronunciation similarities of ``launch'' and ``lunch'' can be considered by the SVG with the strong GPT-2 based PLM, which therefore successfully directed the model to look at the correct entity. In Case 2, although ``abs'' was in the ASR output, the ASR was uncertain about this prediction and ``abbas'' also had a good ASR score. Therefore, while both systems were able to fill the correct slot type, the audio-grounded SVG was able to replace the entity with the one seen in the training set that had similar pronunciation. This also explains why the pipeline-based baseline performed slightly better than the audio-grounded SVG baseline on unseen entities (see the last column of the first two rows in Table \ref{tab:slurpresult1}).

When TCPGen$_\text{ASR}$ was used (Row 3 in the table) with the pipeline system, the main performance boost of the system was attributed to a better WER. The improvements in WER and both F1 scores were mainly on biasing and unseen entities as those were included in the biasing list for TCPGen. Finally, the full KA2G framework achieved higher overall SLU-F1 and Entity-F1 scores compared to the pipeline system and particularly large improvements were observed on few-shot entities and unseen entities. Moreover, due to the benefit of multi-task training, KA2G achieved a lower overall WER: this reflected the fact that the slot-filling task can positively impact the ASR performance. We further noted that the WER on words in the biasing entities was similar to that of the pipeline system: this indicated that the gain with KA2G on low-frequency entities did not only come from improved ASR. 

\subsubsection{Comparison to Baselines from Prior Work} 
The baseline pipeline system using contextual ASR with the overall SLU-F1 score of 79.7 in Table \ref{tab:slurpresult1} already achieved a better performance than the best system proposed in \citet{TCPGenSLU}. This was mainly due to the formulation of slot filling as a sequence generation task. \citet{SLURP2} adopted a sequence labelling approach and reported an overall SLU-F1 score of 78.0 on SLURP (\textit{cf.}, 80.6 reported with KA2G in Table \ref{tab:slurpresult1}). {Note that the work of \citet{SLURP2} used the more powerful WavLM pretrained speech representations \cite{Chen:2022wavlm}, which resulted in a much lower WER of 9\%.} Concerning generation-based slot filling systems, \citet{SLU1} achieved a score of 78.9 using a sequence generator with wav2vec 2.0 as representations. We also found that our pipeline system which as used as the main baseline was usually much better than other pipeline systems reported, as our pipeline adopted the pre-trained GPT2 for slot filling whereas others usually compared to an NLU network with a similar size to the NLU modules in their end-to-end systems \citep{SLURP2,espnetslu}.

\begin{figure}[h]
    \centering
    \includegraphics[width=0.5\linewidth]{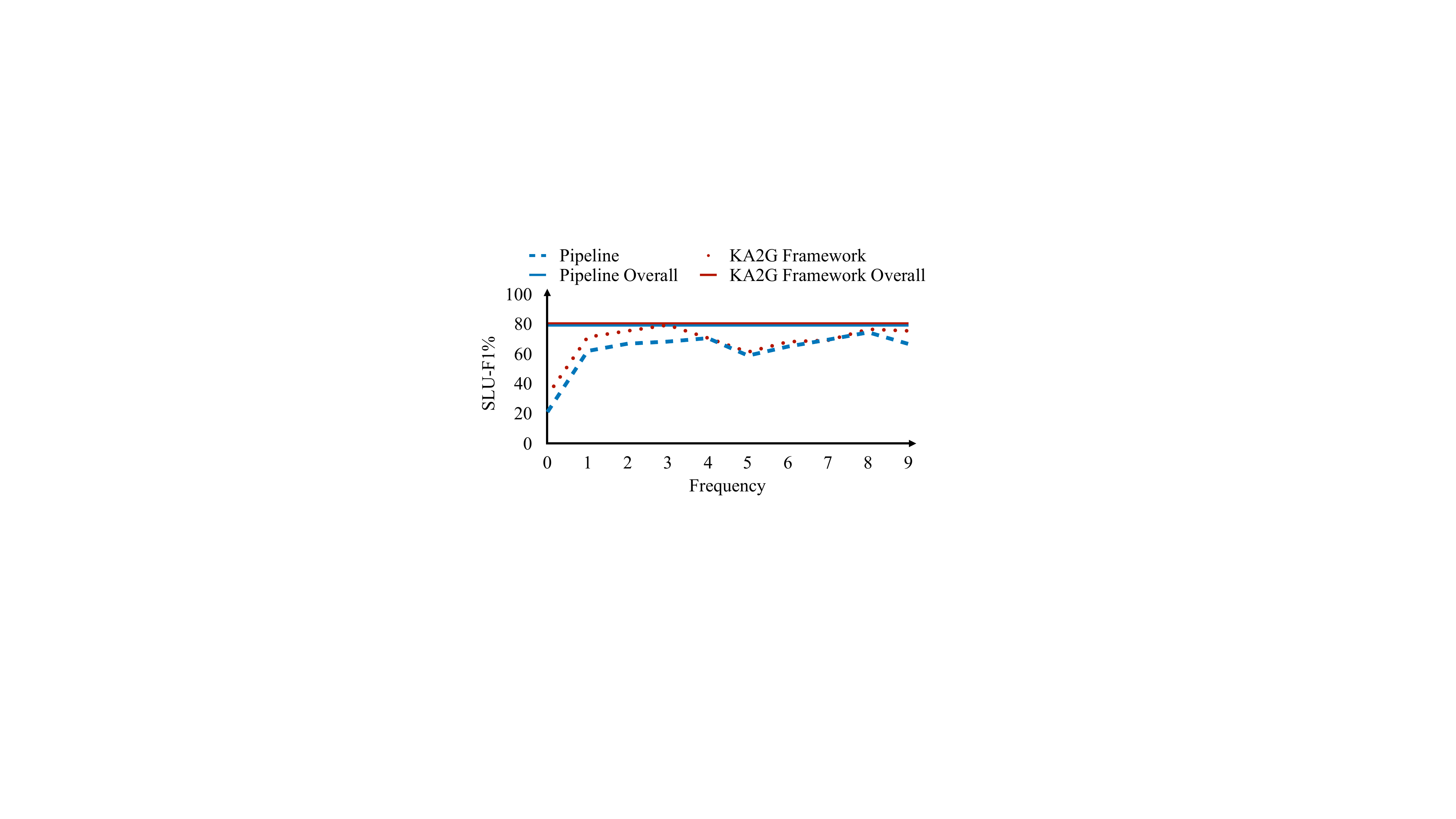}
    \caption{SLU-F1 (\%) over different training set occurrence frequencies for entities in SLURP. \textit{Overall} SLU-F1 scores were also provided as horizontal lines.}
    \label{fig:fewshot}
\end{figure}
\subsubsection{Few-Shot versus Zero-Shot} 
The preliminary comparison of results between few-shot and unseen entities from Table \ref{tab:slurpresult1} showed that, even when only a handful of examples were provided, the model was able to achieve a sizable jump in performance. To provide more insight, a finer-grained frequency bin analysis was conducted on SLURP, as shown in Fig. \ref{fig:fewshot}. The results show that there is a very large increase in SLU-F1 after only providing a single sample for an entity (i.e., moving from zero-shot to one-shot), with SLU-F1 scores increasing from $\sim$30 to $\sim$70: this corroborated the major benefits of few-shot learning over zero-shot learning \cite{Lauscher:2020emnlp}. Fig. \ref{fig:fewshot} again validates that KA2G provides improvements over the baseline system in such low-resource scenarios. 

\subsubsection{Ablation Study}
\begin{table*}[t]
    \centering
    {
    \begin{tabular}{lccc}
    \toprule
    \textbf{System}  & \textbf{Overall} (\%) & 0 $<$ f $<$ 5 (\%) & f $=$ 0 (\%)\\ 
    \cmidrule(lr){2-4}
     KA2G framework & \textbf{80.6 (75.1)} & \textbf{73.8 (73.7)} & \textbf{32.3 (18.6)}\\
      ~~~~  without TCPGen$_\text{SVG}$ & 80.6 (74.7) & 73.6 (73.5) & 27.5 (8.1) \\
      ~~~~  without TCPGen$_\text{ASR}$ & 79.7 (74.3) & 72.5 (72.3) & 27.4 (13.5) \\
      ~~~~  without TCPGen$_\text{SVG}$ and $P^\text{gen}$ input & 80.2 (74.5) & 73.1 (73.0) & 24.4 (7.5) \\
      ~~~~  without TCPGen$_\text{SVG}$ and TCPGen$_\text{ASR}$ & 79.6 (73.9) & 72.2 (71.9) & 20.9 (6.5)\\
    \midrule
    KA2G framework + SVG beam search & \textbf{80.6 (75.2)} & \textbf{74.2 (74.1)} & \textbf{32.4 (18.8)} \\
        ~~~~  without SVG-TCPGen and ASR-TCPGen & 79.6 (73.9) & 72.6 (72.2) & 21.1 (7.2) \\
    \bottomrule
    \end{tabular}
    }%
    \caption{An ablation study on SLURP based on SLU-F1 (Entity-F1 in parentheses). F1 scores were measured for all entities, as well as for few-shot entities and unseen entities. The first row referred to the complete KA2G framework, and each subsequent row represented removing the corresponding components.}
    \label{tab:slurpresult2}
\end{table*}

The results of the ablation experiments for KA2G are given in Table \ref{tab:slurpresult2}, where the last row in the table corresponds to the system which used the audio-grounded SVG (i.e., the second row in Table \ref{tab:slurpresult1}). By removing the TCPGen$_\text{SVG}$, the most significant change was the decrease in performance on unseen entities, especially in terms of Entity-F1. When TCPGen$_\text{SVG}$ was included, the SVG was fully guided by the biasing entities, and hence the model was more likely to predict complete entities. This observation was also found by comparing the system without TCPGen$_\text{ASR}$ to the system without TCPGen$_\text{SVG}$. Moreover, since the WERs for unseen entities were much higher for the 1-best hypothesis, extracting information from alternative hypotheses is much more useful for unseen entities.

While TCPGen$_\text{ASR}$ contributed to the performance improvement, the use of the copy probability $P^\text{gen}$ also contributed to the improvement, especially for rare and unseen entities, as it provided the indication of where knowledge has been used. This was particularly useful when the entity was correctly recognised but SVG was not able to generate it due to not seeing enough examples.

\subsubsection{Impact of Training Data Size} 

The impact of the training data size on the performance of few-shot entities was also analysed. Specifically, the 2.6k utterances which contain few-shot entities were retained while the rest of the training set was sub-sampled. These subsets were then used to train the pipeline system and the full KA2G framework. Unlike other sequence-to-sequence slot-filling frameworks \cite{SLU1} with speech input, the ASR component in KA2G can potentially benefit from training as a standalone module on domain-specific ASR data. Since ASR annotation is usually easier to obtain than SLU annotation, the ASR component in both the pipeline system and KA2G could benefit from the ASR annotation of the full SLURP training data. To this end, two sets of experiments were conducted to investigate the impact of the training data size, where the first experiment trained both the ASR and the SLU on the same selected subset with TCPGen, whereas the second experiment used the full SLURP training set to train the ASR with TCPGen and the selected subset to train the SLU.
\begin{figure*}[t]
    \centering
    \includegraphics[width=0.9\linewidth]{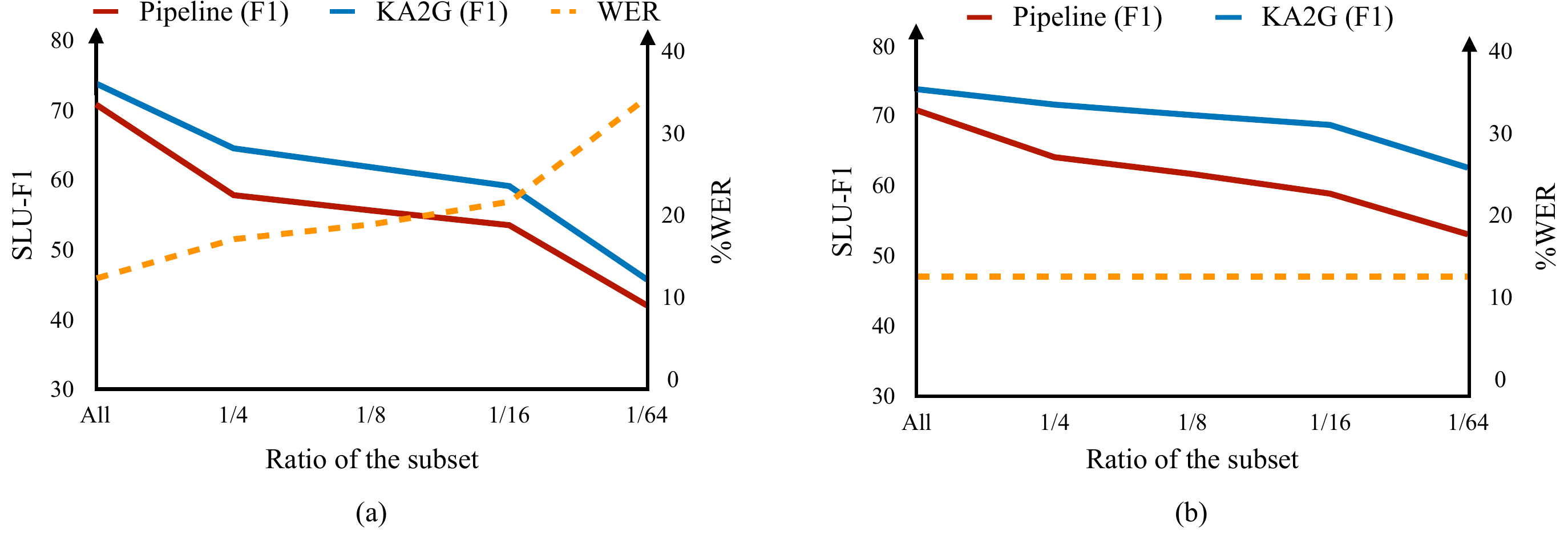}
    \caption{SLU-F1 on few-shot entities when subsampling the part of the training set not containing few-shot entities. (a). The ASR component in the pipeline and KA2G systems were trained on the same subset. (b). The ASR was first trained on the full SLURP training set but SLU was only trained on the selected subset.}
    \label{fig:subset1}
\end{figure*}

The results are shown in Fig. \ref{fig:subset1}. Reducing other training data indeed had a strong impact on SLU-F1 scores of few-shot entities despite the fact that the utterances covering those entities were retained. With the ASR component trained on the full SLURP data (i.e. the second set), the reduction in SLU-F1 became much smaller. The full KA2G again consistently outperformed the baseline system across different training data sizes, with a larger difference when the ASR module was trained on the full data.

\subsubsection{The beam search} 

Beam search can be performed for both ASR and SVG. For ASR, it was found that a higher beam size only yields very marginal improvements in WER while taking significantly more decoding time. From the perspective of the biasing lists for TCPGen$_\text{SVG}$, in the one-best hypothesis, only 50\% of entities can be found. With a beam size of 30, 70\% of entities were covered in the biasing lists for TCPGen$_\text{SVG}$, which was the main source of improvements for TCPGen$_\text{SVG}$. However, this was only improved by 2\% using a beam size of 100, whereas the average size of the biasing lists increased from 10 to 15 which introduced more noise to the biasing lists. Therefore, a beam size of 30 was used for the rest of the experiments. On the SVG side, beam search only provided a marginal improvement, but the time for decoding was 4--5 times longer as the lengths of alternative slot values were in general longer than \texttt{None}.

\subsubsection{Impact of External Language Model (LM) Fusion}
ASR models usually integrate external language information effectively via LM fusion to further boost their performance, hence it is worthwhile studying the effect of using an external LM in the SLU context for both pipeline systems and KA2G. As AED models used in this paper already inherently contained an implicit internal LM that is trained on the SLURP data, an external LM has to contain richer LM information to be effective for ASR. Therefore, a powerful GPT2 PLM finetuned on the text of the SLURP training set was employed. As the modelling unit of GPT2 was different to the AED model, the finetuned GPT2 was used to perform rescoring for the 30-best list from the AED model. For the pipeline system, the re-ranked 1-best hypothesis was used by the text-based SVG, and for KA2G, the hidden states of the 1-best hypothesis were cached and used by the audio-grounded SVG module. The results for both systems are shown in Table \ref{tab:lmeffect}. 
\begin{table}[h]
    \centering
    {
    \begin{tabular}{l cc}
    \toprule
    \textbf{System}  & WER (\%) & SLU-F1 \\ 
    \cmidrule(lr){2-3}
     Pipeline & 12.7 &  79.2 \\
     Pipeline + GPT2 & 12.3 & 79.0 \\
     KA2G framework & 12.1 & \textbf{80.6} \\
     KA2G framework + GPT2 & 11.7 & 80.5 \\
    \bottomrule
    \end{tabular}
    }%
    \caption{Investigation of external LM effects on the pipeline system and KA2G. GPT2, as the external LM, was finetuned on the text of the full SLURP training data. The LM scaling factor was set to 0.3.}
    \label{tab:lmeffect}
\end{table}

Although rescoring with GPT2 further reduced the WER by 0.4\% in absolute value for both systems, it did not improve the SLU performance. In fact, the external LM also tended to reinforce the common correlation in the text which had been well-modelled by the SVG module, and it was the performance on the low-frequent entities that needed to be improved. As a result, external LM integration had a very limited or even negative influence on slot-filling.

\subsubsection{SLURP: Zero-Shot Setup}
\begin{table}[h]
    \centering
    {
    \begin{tabular}{lcc}
    \toprule
    System  & SLU-F1 (\%) & Entity-F1 (\%)\\ 
    \midrule
     Pipeline    & 10.1 & 3.7 \\
     KA2G framework & \textbf{23.7} & \textbf{12.9} \\
        ~~~~ w/o TCPGen & 9.3 & 3.0 \\
    \bottomrule
    \end{tabular}
    }%
    \caption{SLU-F1 and Entity-F1 scores for unseen slots under the zero-shot learning setup on SLURP. Note that w/o TCPGen refers the system that removed both TCPGen$_\text{ASR}$ and TCPGen$_\text{SVG}$.}
    \label{tab:slurpresult3}
\end{table}
The results of the zero-shot setup are summarised in Table \ref{tab:slurpresult3}. The WER for the zero-shot learning test set was 18.0\% for ASR without TCPGen, which decreased to 16.9\% for the KA2G. Compared to the pipeline system, KA2G achieved worthwhile improvements both in SLU-F1 and Entity-F1 scores. Therefore, KA2G provided an effective way of leveraging knowledge for zero-shot slot filling by bridging the SVG and ASR alternative hypotheses via a neural shortcut provided by TCPGen. Moreover, while zero-shot slot filling in \citet{TCPGenSLU} relied on manually tuned hyper-parameters, KA2G removed the requirement on TCPGen-related hyperparameter tuning during inference, which also improved the robustness and reliability of KA2G for zero-shot slot-filling.

\subsection{Experiments on CONCIERGE}
\vspace{0.3cm}
This section discusses the results of the CONCIERGE data under single-turn and multi-turn evaluation metrics.

\subsubsection{Single-Turn Evaluation}
The proposed KA2G framework was also validated on a real-world use case with the CONCIERGE dataset. The WER on this test set was $\sim$35\% due to limited audio data for training, which made the slot-filling task with speech input on CONCIERGE even more challenging. Single-turn evaluation was first performed, with the same metrics as used with SLURP. The results are provided in Table \ref{tab:polyresult1}. 

In the challenging setup, external knowledge plays an even more important role, which led to much larger performance improvements with the KA2G framework. As with SLURP, having a few examples of entities yielded much better performance than zero-shot learning. Clearly, providing a handful of examples in the training set resulted in very large performance improvements. This again corroborated the fact that although zero-shot learning is an attractive research problem, few-shot learning is often more pragmatic for industrial applications, and this is also the case when using generative systems such as KA2G.

\begin{table}[h]
    \centering
    {
    \begin{tabular}{lcc}
    \toprule
    System  & SLU-F1 (\%) & Entity-F1 (\%)\\ 
    \midrule
     Pipeline    & 37.2 & 26.9 \\
     KA2G framework & \textbf{56.6} & \textbf{42.5} \\
        ~~~~ w/o TCPGen & 41.9 & 29.3 \\
    \midrule
    Pipeline zero-shot & 1.6 & 1.6 \\
    KA2G zero-shot & 12.4 & 2.6 \\
    \bottomrule
    \end{tabular}
    }%
    \caption{SLU-F1 and Entity-F1 scores on CONCIERGE. Zero-shot results were obtained by removing dialogues containing test set entities from training.}
    \label{tab:polyresult1}
\end{table}

\begin{table}[h]
    \centering
    {
    \begin{tabular}{lc}
    \toprule
    System  & JGA (\%) \\ 
    \midrule
     Pipeline    & 21.0 \\
     KA2G framework & \textbf{41.5} \\
        ~~~~ w/o TCPGen & 24.3 \\
    \bottomrule
    \end{tabular}
    }%
    \caption{JGA on the CONCIERGE dataset with multi-turn dialogue state tracking.}
    \label{tab:polyresult2}
\end{table}

\subsubsection{Multi-Turn Evaluation}
For multi-turn experiments, entity mapping was applied in order to group different expressions of the same entity together. Further, the ASR 1-best hypothesis from the history of user inputs was included as input to the PLM. The JGA scores are summarised in Table \ref{tab:polyresult2}. There were large improvements with the full KA2G, which were mostly due to the use of TCPGen$_\text{SVG}$. As JGA was more relevant to Entity-F1, and the SVG-TCPGen module in the KA2G framework provided particular benefits to Entity-F1, the KA2G framework resulted in a clear improvement in JGA.

\section{Conclusions}
\label{sec:conclusion}

A novel knowledge-aware audio-grounded generative slot-filling framework for speech-based ToD, called KA2G has been proposed. The framework is especially suited for low-resource slot-filling tasks and for handling rare and unseen entities/values. KA2G comprises an audio-grounded SVG, together with two TCPGen components. 
The first TCPGen integrates knowledge from an external knowledge base containing possible entities for all slots into the ASR module, while the second TCPGen exploits entities found in alternative ASR hypotheses. A comprehensive evaluation has been performed on two different datasets with speech input: i) single-turn SLURP data and ii) multi-turn CONCIERGE data obtained from a commercial ToD system. The usefulness of KA2G has been experimentally validated on both datasets, with clear performance gains over current state-of-the-art systems. KA2G was especially useful in few-shot and zero-shot setups. 

KA2G, as a prompt-based SLU framework with speech input, also possesses potential for future investigation with the advent of large language models (LLMs) and prompt-based generative AI. 
We believe KA2G can serve as a promising speech front-end for LLMs. The explicit knowledge integration component that allows dynamic contextual knowledge to be incorporated in a prompt-based system may also potentially benefit the performance of LLMs on factual and domain-specific enquiries.

\bibliographystyle{cas-model2-names}

\bibliography{cas-refs}


\end{document}